# From Density to Biopsy Decisions and Malignancy Prediction: A Benchmark Study of Multimodal Large Language Models Against Radiologists in Digital and Contrast-Enhanced Mammography

Ali Abbasian Ardakani[1]; Afshin Mohammadi[2]; Taha Yusuf Kuzan[3]; Beyza Nur Kuzan[4]; Alisa Mohebbi[1]; Masume Behruzi[5]; Hamid Khorshidi[6]; Ashkan Ghorbani[7]; Elham Asadiara[8]; Zeinab Khorshidi Lotfi[9]; Ansar Rahman[1]; Nedim Christoph Beste[10]; U. Rajendra Acharya[11, 12]; Sepideh Hatamikia[1, 13*]

1 Department of Medicine, Faculty of Medicine and Dentistry, Danube Private University, Krems an der Donau, Austria

2 Department of Radiology, Faculty of Medicine, Urmia University of Medical Science, Urmia, Iran

3 Department of Radiology, Yeditepe University, Istanbul, Türkiye

4 Kartal Dr. LÜtfi Kırdar City Hospital, Istanbul, Türkiye

5 Department of Anatomical Sciences, School of Medicine, Iran University of Medical Sciences, Tehran, Iran

6 Department of Information Engineering, University of Padova, Padova, Italy

7 Department of Bioengineering, Bahçeşehir University, Istanbul, Turkey

8 Mooney's Bay Pain Clinic, Ottawa, Ontario, Canada

9 School of Computing, Faculty of Social Sciences and Technology, Arden University, Berlin, Germany

10 Institute for Diagnostic and Interventional Radiology, University Hospital Cologne, Cologne, Germany

11 School of Mathematics, Physics and Computing, University of Southern Queensland, Springfield, Australia

12 Centre for Health Research, University of Southern Queensland, Springfield, Australia

13 Austrian Center for Medical Innovation and Technology, Wiener Neustadt, Austria

**∗ Corresponding author:**

**Sepideh Hatamikia, E-mail:** Sepideh.hatamikia@DP-Uni.ac.at

# From Density to Biopsy Decisions and Malignancy Prediction: A Benchmark Study of Multimodal Large Language Models Against Radiologists in Digital and Contrast-Enhanced Mammography

**Abstract**

**Purpose:** To compare four multimodal large language models (MLLMs) with radiologists of varying expertise in breast density assessment, BI-RADS assessment, biopsy candidacy determination, and continuous malignancy probability estimation using digital mammography (DM) and contrast-enhanced mammography (CEM).

**Methods:** This study included 179 women with paired DM/CEM examinations and reference standards. Four MLLMs (ChatGPT-5.2, Gemini-3.1 Pro, Sonnet-4.6, Muse Spark) interpreted images with and without masks; three radiologists interpreted non-masked images.

**Results:** For binary density classification on DM, radiologist accuracies ranged from 55.81% to 78.60%, exceeding most MLLM values (62.33%-71.63%), while masks added limited benefit. Five-category BI-RADS accuracies were higher for radiologists on DM (56.74-67.44%) and CEM (62.33-82.79%) compared with MLLMs (DM 31.16-45.12%; CEM 40.00-55.81%). Binary biopsy-candidacy accuracies were likewise higher for radiologists (DM 85.12-89.77%; CEM 86.98-92.09%) than for MLLMs (DM 61.39-75.35%; CEM 69.30-82.79%), although CEM improved performance across all readers. Lesion masks substantially improved MLLM continuous malignancy-probability accuracies from 64.65%-71.63% to 72.56-78.60% on DM and from 67.91%-77.21% to 72.56%-81.86% on CEM, approaching radiologist ranges (DM 63.72-82.79%; CEM 81.86-88.84%). The corresponding AUCs for the top masked models overlapped those of the human readers. Overall, Muse Spark, followed by Sonnet-4.6, demonstrated the strongest performance among the MLLMs across domains.

**Conclusion:** Radiologists generally outperformed MLLMs in categorical tasks, while selected masked models approached human performance for continuous malignancy probability estimation, suggesting a potential adjunctive role.

**Keywords:** Contrast-enhanced mammography; Digital mammography; Large language models; Artificial intelligence; BI-RADS.

# 1. Introduction

Contrast-enhanced mammography (CEM) is now an important adjunct to conventional digital mammography (DM). It combines dual-energy imaging with intravenous iodinated contrast to show lesion neovascularity that may be missed on standard low-energy mammograms (1-4). DM is still the main modality for initial workup, but its sensitivity is limited in women with dense breast tissue, where overlapping fibroglandular parenchyma can hide malignancy (5, 6). CEM addresses this limitation by creating a low-energy image similar to DM and an additional contrast-enhanced image that highlights areas of abnormal enhancement, thereby making it easier to evaluate lesions. Precise assessment of breast density on mammography, accurate BI-RADS categorization, and reliable quantitative estimation of malignancy probability are key tasks that guide clinical decisions. However, these assessments remain subject to inter-reader variability (7, 8). Differences in radiologist experience, especially between fellowship-trained breast radiologists and general radiologists, may increase this variability across both DM and CEM examinations.

Multimodal large language models (MLLMs) have recently attracted attention as potential decision-support tools in breast imaging (9-13). MLLMs may be prompted to review DM and CEM images after they are uploaded, and in response, can be asked to generate structured reports, assign density categories, determine BI-RADS scores, suggest biopsy recommendations, and estimate cancer risk. They may be able to process both lesion-segmented images and non-segmented mammograms when these are uploaded and the models are prompted accordingly; in principle, this could help reduce interpretive variability and supply rapid second opinions, particularly in settings with limited access to fellowship-trained breast imaging expertise. However, while early studies have examined MLLM performance on DM (9-12, 14) research on CEM as well as systematic comparisons that evaluate multiple MLLMs against radiologists of differing experience levels remain limited.

Therefore, there is a clear gap in the literature regarding systematic, head-to-head comparisons of MLLMs and human readers with different levels of experience when interpreting both DM and CEM using segmented versus non-segmented conditions. To address this gap, this study aimed to evaluate the diagnostic performance of four MLLMs (ChatGPT, Gemini, Sonnet, and Muse Spark) and three radiologists for four main clinical outcomes: (1) determining breast density category, (2) assigning a BI-RADS score, (3) determining candidacy for biopsy, and (4) quantitatively estimating malignancy probability (%). We compared DM and CEM using both segmented inputs and non-segmented images.

# 2. Materials & Methods

### 2.1. Study Design & Eligibility Criteria

This retrospective study evaluates the interpretive performance of four MLLMs against that of three radiologists with varying levels of experience. The study followed the Standards for Reporting of Diagnostic Accuracy Studies (STARD) and Transparent Reporting of a Multivariable

Model for Individual Prognosis or Diagnosis (TRIPOD)-LLM guidelines. The imaging data were sourced from the CDD-CESM (15). Institutional review board approval was obtained by the data curators at the National Cancer Institute, and written informed consent from all patients. Because the present study used only fully anonymized data, the local institutional review board waived the requirement for further approval. Details of the inclusion and exclusion criteria are provided in Fig. 1A.

### 2.2 Image Preparation & Presentation Conditions

Both craniocaudal and mediolateral oblique views from each examination were prepared and given to all readers (the four MLLMs and the three radiologists). This ensured that everyone had the full standard set of mammograms for every case. Interpretation was performed in two separate sessions to eliminate memory bias. In the first session, only the DM images were presented to each MLLM and radiologist. Following a mandatory one-month washout period, the identical cohort was re-presented to the radiologists, with CEM images added to the DMs. In contrast, no washout period was applied for the MLLMs, as the models were configured not to retain prior interpretations.

In each modality session, the lesion-presentation conditions depended on the type of reader. For the four MLLMs, two parallel sub-conditions were used: (1) non-segmented full-field images, where the whole mammographic field of view was provided without any mask, and (2) segmented images, where a manual pixel-level segmentation mask of the lesion(s) was provided along with the mammograms. This dual-input strategy for the MLLMs was designed to test the hypothesis that providing explicit lesion segmentation would improve model performance. In contrast, the three radiologists interpreted only the raw (non-segmented) versions of each image. Fig. 1B illustrates the strategy used for the MLLMs and radiologists. Detailed imaging acquisition parameters are provided in the Supplementary Material.

### 2.3. Reference Standard Construction

For each examination, the reference labels were defined through blinded evaluation by two breast radiologists, as provided in the CDD-CESM dataset (15). Breast density was assessed on DM images using the four-category ACR BI-RADS density classification. For secondary analyses involving binary classification, the four categories were combined into low density (A + B) versus high density (C + D). The ground-truth BI-RADS score was also determined by the same radiologists. For biopsy candidacy determination, BI-RADS categories 1-3 were labeled as “not candid” (no biopsy needed), while categories 4-5 were labeled as “candid” (biopsy needed).

Malignancy status was determined by histopathology or by documented long-term imaging follow-up, which served as the reference standard. We used this reference to compare the cancer probability scores by the MLLMs and the radiologists using receiver-operating-characteristic (ROC) analysis.

### 2.4. MLLM Selection and Configuration

Four commercially available MLLMs that accept both image and text inputs were evaluated: ChatGPT-5.2, Gemini-3.1 Pro, Sonnet-4.6, and Muse Spark. These particular versions represented the latest versions available at the time the study was conducted (early 2026). All models were instructed to act as fellowship-trained breast imaging radiologists and were given a concise reference document summarizing the ACR BI-RADS guidelines for DM and CEM. For each exam, we used structured prompts to ask the models to return the BI-RADS density category, overall BI-RADS assessment, biopsy-candidacy decision, a continuous malignancy probability estimation (0-100), and final diagnostic decision. The full prompt texts and the exact interaction protocol are provided in Supplementary Material.

### 2.5. Human Reader Protocol

Three board-certified radiologists interpreted every image (with 5, 7, and 9 years of post-residency experience in mammography, respectively). Each radiologist was provided with a standard electronic form that matched the output template required of the MLLMs. The radiologists were provided with the same concise ACR BI-RADS reference document given to the MLLMs to ensure consistency and comparability between the human and model assessments. If multiple lesions were available per-patient, only lesion with the highest ACR BI-RADS score was assessed in all examinations. The readers were blinded to the reference-standard diagnoses, to each other's interpretations, and to all MLLM outputs.

### 2.6. Statistical Analysis

The analyses were performed using R version 4.4.2. Because the unit of analysis was the breast rather than the patient, statistical methods that account for clustering of bilateral breasts within individual patients were used throughout. For each aim, Gwet's AC1, area under the curve (AUC), sensitivity, specificity, and accuracy were calculated. For breast density, sensitivity was the proportion of reference-standard high-density (BI-RADS C+D) examinations correctly classified as high density, and specificity was the proportion of reference-standard low-density (BI-RADS A+B) examinations correctly classified as low density. For the biopsy-candidacy aim, sensitivity was the proportion of reference-standard biopsy-candidate lesions (BI-RADS 4-5) correctly classified as candidates, and specificity was the proportion of reference-standard non-candidate lesions (BI-RADS 1-3) correctly classified as non-candidates. For the malignancy classification goal, sensitivity was the proportion of reference-standard malignant lesions correctly classified as malignant, and specificity was the proportion of reference-standard benign lesions correctly classified as benign. Comparisons of AUCs between readers were performed with the DeLong's method for clustered data. All statistical tests were two-sided, and a p-value $<0.05$ was considered significant.

## 3. Results

The final cohort consisted of 179 women who met all inclusion criteria and had complete DM and CEM examinations, manual pixel-level lesion segmentations, and a gold standard (Fig. 1a). All seven readers (four MLLMs and three radiologists with different levels of experience) interpreted every examination.

### 3.1. Breast-density Assessment

Performance metrics for four-category and binary (low vs. high) breast-density classification on DM are presented in Table 1 and illustrated in Fig. 2a. For ACR BI-RADS density classification, radiologists 1, 2, and 3 achieved accuracies of 36.74%, 67.91%, and 59.07%, respectively, while the MLLMs' accuracies ranged from 51.16% to 57.21%. The Sonnet with-mask and Muse Spark without-mask methods outperformed the other MLLMs, with accuracies of 57.21% and 56.74%, respectively, which were numerically better than that of radiologist 1, similar to that of radiologist 3, but lower than that of radiologist 2.

When density was evaluated as a binary low-versus-high classification, the three radiologists had accuracies ranging from 55.81% to 78.60%. On the other hand, the four MLLMs operating without lesion masks yielded accuracies between 63.26% and 71.63%. When these models were given segmentation masks, their accuracies ranged from 62.33% to 68.37%.

The radiologists generally performed better than both the without-mask and with-mask MLLM groups. Segmentation provided no additional performance benefit, as accuracies for the MLLM cohort remained essentially unchanged or slightly lower (for Sonnet) when masks were supplied. Amongst the MLLMs, Muse Spark with mask and Sonnet without mask provided the best performance, significantly better than radiologist 1, similar to radiologist 2, but significantly lower than radiologist 3. Although no significant difference was seen between the two MLLMs, Muse Spark showed better overall performance according to AUC (0.706 vs. 0.642) reflecting balanced performance between low and high breast density (Table 1, Fig. 4a).

### 3.2. BI-RADS Assessment & Biopsy-candidacy Determination on DM

Full results for multi-category BI-RADS score assignment and binary biopsy-candidacy determination on DM are available in Table 2 and Fig. 2b. For the five-category BI-RADS assessment, the radiologists achieved higher accuracies, ranging from 56.74% to 67.44%. In comparison, the without-mask MLLM group had accuracies between 35.35% and 40.93%, while the with-mask MLLM group yielded accuracies between 31.16% and 45.12%. In this regard, the ChatGPT with-mask and Muse Spark with-mask methods outperformed the other MLLMs, with accuracies of 45.12% and 40.93%, respectively.

For the clinically important binary distinction between non-candidates (BI-RADS 1-3) and candidates (BI-RADS 4-5), radiologists accuracies ranged from 85.12% to 89.77%. The without-mask MLLM group had accuracies between 64.17% and 71.16%, while the with-mask MLLM group had accuracies between 61.39% and 75.35%. The best performance was seen for the Muse Spark with-mask method with 75.35% accuracy. No significant differences were seen between the three radiologists. However, all radiologists significantly outperformed every MLLM ($P<0.05$). Adding a segmentation mask did not significantly improve the accuracy of the MLLMs. Types I and II errors of the best-performing MLLM, Muse Spark, and three radiologists are shown in Fig. 4b.

### 3.3 BI-RADS Assessment & Biopsy-candidacy Determination on CEM

Complete performance metrics for both the five-category BI-RADS score assessment and the binary biopsy-candidacy determination on CEM images are presented in Table 3 and Fig 3a. For the five-category (BI-RADS 1-5) assessment, the radiologists achieved accuracies between 62.33% and 82.79%. On the other hand, the without-mask MLLM group achieved accuracies between 40.00% and 48.84%, while the with-mask MLLM group reached 42.33%-55.81%. Compared with the DM results, all reader classes improved; radiologists' five-category accuracies rose from the DM range of 56.74%-67.44% to 62.33%-82.79%, the without-mask MLLM group's accuracies rose from 35.35%-40.93% to 40.00%-48.84%, and the with-mask MLLM group's accuracies rose from 31.16%-45.12% to 42.33%-55.81%. Muse Spark performed the best in both without- and with-mask MLLM groups, with accuracies of 48.84% and 55.81%, respectively.

Regarding the biopsy candidacy decision, radiologists' accuracies ranged from 86.98% to 92.09%. The without-mask MLLM group's accuracies were 69.30%-81.39% and the with-mask MLLM group's accuracies reached 73.95%-82.79%. The CEM accuracies for the binary classification task were also uniformly higher than the corresponding DM accuracies (radiologists, from 85.12%-89.77% to 86.98%-92.09%; without-mask MLLMs, from 64.17%-71.16% to 69.30%-81.39%; and with-mask MLLMs, from 61.39%-75.35% to 73.95%-82.79%). Statistical analysis revealed that radiologists performed significantly better than most MLLMs. However, the absolute accuracy gap narrowed compared with that of the best-performing MLLM (Muse Spark). Muse Spark (both without- and with-mask methods) performed significantly better than most other MLLMs, worse than radiologists 1 and 2, and similarly to radiologist 3. However, the Muse Spark with-mask method showed higher sensitivity and lower specificity than the without-mask method (Table 3 and Fig. 4d). No significant differences were seen between the three radiologists. Adding a mask did not significantly improve accuracy for any of the MLLMs.

### 3.4. Malignancy Classification on DM

Figures 2c, and 5 and Table 4 show the full results for the malignancy prediction task. Using the continuous malignancy-probability estimates, radiologists' accuracies and AUCs for benign-versus-malignant discrimination ranged from 63.72% to 82.79%, and from 0.805 to 0.883, respectively. The without-mask MLLM group achieved accuracies and AUCs of 64.65%-71.63% and 0.656-0.728, respectively, while the with-mask MLLM group achieved accuracies and AUCs of 72.56%-78.60% and 0.759-0.874, respectively.

Mask usage significantly increased the performance of ChatGPT, Sonnet, and Muse Spark, but did not show significant improvement for Gemini. Muse Spark and Sonnet (both with the masked method) showed the best performance among the MLLMs, with accuracies of 78.60% and 77.21%, and AUCs of 0.834 and 0.874, respectively. Muse Spark indicated a higher false-negative rate (Type II error) and a lower false-positive rate (Type I error) than Sonnet (Fig. 4c). The performance of the masked versions of Muse Spark and Sonnet was similar to that of all radiologists ($P>0.05$).

Figure 5-left shows the distribution of malignancy probability for the MLLMs and radiologists. The radiologists' predictions show a clear separation between benign and malignant cases, giving low scores to benign lesions and high scores to malignant ones. For the MLLMs, the greatest separation was seen in the best-performing models, Muse Spark and Sonnet, suggesting potential utility as adjunctive tools for malignant lesion identification, pending further validation.

### 3.5. Malignancy Classification on CEM

Fig. 3b, and Table 4 show the full results for the malignancy prediction task. On CEM images, the radiologists attained accuracies of 81.86%-88.84% and AUCs of 0.886-0.941. These values represented clear significant improvements (Fig. 3c) relative to their performance on DM (63.72%-82.79%, and 0.805-0.883). Without-mask MLLMs reached accuracies of 67.91%-77.21% and AUCs of 0.720-0.826 on CEM, higher than their performance on DM (64.65%-71.63%, 0.656-0.728). In this regard, adding CEM to DM images significantly improved the performance of ChatGPT, Sonnet, and Muse Spark using the without-mask method (Fig. 3c, Table S1). With-mask MLLMs achieved accuracies of 72.56%-81.86% and AUCs of 0.761-0.875 on CEM, similar to their performance on DM (72.56%-78.60%. and 0.759-0.874, $P>0.05$).

Mask usage significantly improved the performance of Muse Spark and Sonnet. The performance of the masked versions of Muse Spark and Sonnet and that of radiologists 1 and 3 was similar but significantly lower than that of radiologist 2 (Fig. 3b). However, Muse Spark showed a better balance between sensitivity and specificity than Sonnet (Fig. 4e). Sankey flow diagrams of all MLLMs for each task are provided in Fig. S1-S8 in the Supplementary Material.

Figure 5-right shows the distribution of malignancy probability for the MLLMs and radiologists. On CEM, the separation between benign and malignant lesions improved for all readers. The MLLMs' probabilities became closer to the radiologists' probabilities. Muse Spark and Sonnet showed the best separation among the MLLMs, approaching radiologists' performance on CEM.

Fig. 6 shows clinical examples of the best MLLMs (masked versions of Muse Spark and Sonnet) alongside radiologists' outputs. In the first and second cases, both models helped radiologists diagnose a malignant lesion using both DM and CEM (BI-RADS 3 on DM and 5 on CEM), while a benign mimic lesion on DM was not missed (Fig. 6a-b). In the third case, all readers correctly identified a benign lesion with a BI-RADS score of 2 (Fig. 6c). In the fourth case, Muse Spark helped radiologists diagnose a benign lesion (BI-RADS 4 on DM and 3 on CEM), meaning that a malignant mimic lesion would not undergo biopsy (Fig. 6d).

## 4. Discussion

This study is the first direct comparison of MLLMs against radiologists with varying levels of experience in interpreting CEM in parallel with DM. By examining four MLLMs under both segmented and non-segmented conditions and comparing them with three human readers with varying levels of experience, the study provides a framework for assessing the potential complementary role of these models in breast imaging. Overall, radiologists were still better at categorizing breast density and assigning BI-RADS categories and biopsy candidacy. However, the gap narrowed when continuous malignancy probability estimates were examined, especially when the models were given lesion masks. Among the MLLMs, Muse Spark, followed by Sonnet-4.6, consistently indicated the best results across all tasks. Their performance approached or matched that of the less-experienced radiologists in several tasks.

It is clinically important that all MLLMs struggled with the four-category density classification and even with the simpler binary low-versus-high task. Breast density is an important factor in mammography, as it affects both the sensitivity of the examination and a woman's lifetime risk (16). Inter-reader variability in density assignment has long been known as a source of inconsistent supplemental-screening recommendations and patient counseling (17). The fact that MLLMs did not improve when explicit lesion boundaries were provided suggests that their density estimation depends less on focal lesion morphology than on the global pattern of fibroglandular distribution across the entire breast volume. Radiologists seem to integrate this spatial information more easily, drawing on years of experience and pattern recognition, which current vision-language architectures may still not do perfectly. The observation that Muse Spark yielded density metrics closest to the radiologists raises the possibility that certain model families might already encode more reliable textural and distributional patterns. In clinical practice, where density reporting directly affects sensitivity and the offer of supplemental ultrasound or MRI, such residual discrepancy highlights the continued necessity of human oversight. Previous studies have examined the performance of MLLMs in determining breast density. In this regard, Sanli et al. demonstrated that XrayGPT (the paid version of ChatGPT) could identify low and high breast density with an accuracy of 82.5% (18). Likewise, Li et al. found that ChatGPT-5, ChatGPT-5-mini, ChatGPT-5-nano, and ChatGPT-4o were able to distinguish breast density categories with accuracies of 56.8%, 34.3%, 24.8%, and 24.3%, respectively (19). Another study showed accuracies of 30.2% and 26.4% for ChatGPT-4o and Claude-3.5, respectively (20). On the other hand, Zhu et al. tested their model, LLaVA-Mammo, on DM images and achieved 66.8% accuracy (21). These previous findings are consistent with our results, in which Sonnet-4.6 without-mask and Muse Spark with-mask achieved the best performance, reaching 57.21% and 56.74% accuracies for four-class and 71.63% and 67.91% accuracies for two-class breast density classification, respectively.

Regarding BI-RADS categorization and the downstream decision of biopsy candidacy, the superiority of the radiologists was more pronounced on DM and remained evident even when recombined contrast-enhanced images were introduced. The five-category BI-RADS scale requires not only the detection of an abnormality but also the characterization of the morphologic descriptors and, in the case of CEM, kinetic and enhancement patterns. Even when given the ACR lexicon, MLLMs appear to struggle with the ordinal granularity of this scale. Binary candidacy decisions showed better results, consistent with the clinical reality that the most consequential threshold is the distinction between categories of “probably benign or less” and “suspicious or highly suggestive”. However, the persistence of statistically significant differences between MLLMs and radiologists in most pairwise comparisons indicated that current models have not yet internalized the full interpretive hierarchy that experienced readers apply. Despite this, Muse Spark again led the MLLMs. In the assessment of BI-RADS scores, the accuracy of MLLMs has been reported only for DM images and differs across previous studies. The reported accuracy rates on DM images were 66.2% for ChatGPT-4 (22); 28.5% for ChatGPT-4V (23); 23.7% and 28.4% (19), 26.4% (20), 9.6% (24) and 66.2% (22) for ChatGPT-4o; 36.9% and 69.3% for ChatGPT-5 (19); 28.1% and 43.6% for ChatGPT-5-mini (19); 17.6% and 20.2% for ChatGPT-5-nano (19); 18.7% for Claude-3.5 (20); 62.5% for XrayGPT (18); and 32.3% for LLaVA-Mammo (21). However, only Kustkun et al. examined the ability of an MLLM, ChatGPT-4o, to determine whether breast lesions were candidates for biopsy and reported 18.2% accuracy (24). In addition, only two studies compared the performance of MLLMs with that of a radiologist, which showed higher accuracy for radiologists than for ChatGPT-5.2-Thinking (72.0% vs. 62.0%) (25) and low agreement between a radiologist and ChatGPT-4o (kappa=0.017) (24). In our study, this aim was also investigated, and our results demonstrated that Muse Spark with mask could identify biopsy-candidate lesions with reasonable performance, particularly on CEM, achieving 82.79% accuracy, which was comparable to that of one radiologist ($P > 0.05$).

Regarding the assessment of malignancy, previous studies have only reported the performance of MLLMs on DM images. In this regard, ChatGPT-5 reached accuracies of 52.8%, 35%, 55.0%, and 58.2%, ChatGPT-5-mini showed accuracies of 47.3%, 40.0%, 43.6%, and 43.5%, ChatGPT-5-nano had accuracies of 47.8%, 21.5%, 52.7%, and 30.9%, and ChatGPT-4o reached accuracies of 42.5%, 30.4%, 48.5% and 40% on four datasets, respectively. This was the only study that compared the performance of a radiologist with that of MLLMs; the radiologist’s accuracy was 88.9% which was higher than that of the MLLMs (19). On the other hand, Chen et al. (26) and Zhu et al. (21) achieved accuracies of 82.3% and 46.3% for Mammo-CLIP and LLaVA-Mammo models, respectively. Furthermore, Ra et al. demonstrated that LLaMA2 could help improve machine-learning models by 1.8% to 6.7% in identifying malignancy (27). A distinctive clinical contribution of this study is examining MLLMs and radiologists on both DM and CEM images, as well as evaluating continuous, point-estimate malignancy probabilities on a 0-100% scale, rather than the traditional range-based probability intervals within BI-RADS categories. To our knowledge, this is the first systematic assessment of such quantitative outputs from both MLLMs across two settings (with and without masks) and radiologists with different levels of experience using DM and CEM. The addition of manual pixel-level masks improved the MLLMs’ performance, elevating some of them to a range statistically comparable to that of human readers across both modalities. Muse Spark and Sonnet-4.6 with masks reached accuracy and AUC values similar to those of the radiologists in the malignancy prediction task. This suggests the possibility that carefully conditioned MLLMs could serve as suggestive second readers for quantitative risk stratification, particularly in settings where access to subspecialty expertise is limited.

Comparing the DM and CEM results for breast lesion malignancy prediction reveals several valuable insights. Adding contrast-enhanced images to DM significantly improved the performance of all three radiologists. A similar trend was observed in most MLLMs without masks. This shows that adding DM to CEM helped these models develop a more comprehensive understanding of breast tissue and better differentiate between malignant and benign lesions. However, adding CEM images did not provide added value for MLLMs with masks and did not significantly improve their performance (Fig. 3c). This finding suggests that when lesion masks are provided on DM, MLLMs can focus on lesions, filter out the complex background parenchyma, and extract enriched information from the lesions, including morphological and textural features as well as hemodynamic information (which is provided by CEM), that cannot be perceived by the radiologists' unaided eyes on DM images alone. This approach allows the MLLMs to reach reliable diagnostic conclusions without the need for CEM. Therefore, they can be potentially considered effective decision-support tools for radiologists in routine practice without CEM technology access, though prospective validation is needed (see cases a-b in Fig. 6).

Comparing our results with the literature reveals that this is the first study to comprehensively examine the performance of MLLMs and radiologists across all available decision tasks in routine clinical practice: breast density classification, BI-RADS score evaluation, biopsy-candidate lesions identification, and malignancy prediction (Table 5). In addition, this is the first study to include CEM alongside DM images to evaluate MLLMs performance on CEM and assess their ability to analyze multimodal data (DM+CEM). Likewise, we also examined the effect of lesion masks on MLLMs to determine whether the masks could help them focus on predefined regions and improve their performance. We reported these results together with the performance of three radiologists to provide a benchmark for future studies.

Several limitations of the present study merit careful consideration when interpreting its findings. First, the retrospective reliance on a single-center cohort may restrict the generalizability of the results to more diverse patient populations, imaging protocols, and institutional practices encountered in everyday clinical work. Second, images were supplied exclusively in JPEG format rather than native DICOM or full-resolution PACS data, raising the possibility that subtle contrast gradients, microcalcification morphology, or faint enhancement patterns critical to expert interpretation were attenuated. Third, each MLLM was ran only once per case, and the outputs were not averaged or aggregated across repeated runs. Because MLLMs may slightly produce different responses when the same case is evaluated multiple times, the reproducibility of the reported results may be limited. Fourth, CDD-CESM is a publicly available dataset, so we could not determine whether the commercial MLLMs were exposed to it or similar cases during training. Therefore, possible data contamination cannot be excluded and may have influenced the reported performance. Finally, the inherently black-box nature of the models precludes granular feature-level analysis of discordant cases, limiting the radiologist's ability to understand, trust, or systematically correct systematic errors.

# 5. Conclusion

Findings suggest that radiologists may retain superiority over current MLLMs in density categorization and BI-RADS assessment across DM and CEM. Muse Spark, followed by Sonnet-4.6, appeared more robust among the models. Lesion masks seemed to enhance continuous malignancy-probability estimation, permitting selected systems to approach radiologist performance. These results point toward a potential adjunctive role rather than independent deployment, warranting further prospective evaluation.

**Acknowledgements**

This research is funded by EU Structural Fund IBW/EFRE and the Lower Austrian Economic and Tourism Fund (Project number: WST3-F-5035462/005-2024).

**Table 1.** Performance of MLLMS and radiologists in identifying breast lesion density in digital mammography.

| Method | Condition | Predicted Class | True Class A | B | C | D | Gwet's AC | Acc (%) | Predicted Class | True Class Low Density | High Density | Gwet's AC | Sen (%) | Spec (%) | Acc (%) | AUC (95% CI) |
|---|---|---|---|---|---|---|---|---|---|---|---|---|---|---|---|---|
| ChatGPT-5.2 | Without Mask | **A** | - | - | - | - | 0.870 | 55.81 | **Low Density** | 15 | 3 | 0.512 | 97.78 | 18.75 | 68.37 | 0.583 (0.501, 0.664) |
| | | **B** | 2 | 13 | 3 | 0 | | | | | | | | | | |
| | | **C** | 1 | 64 | 103 | 15 | | | **High Density** | 65 | 132 | | | | | |
| | | **D** | 0 | 0 | 10 | 4 | | | | | | | | | | |
| | With Mask | **A** | - | - | - | - | 0.866 | 56.28 | **Low Density** | 15 | 3 | 0.512 | 97.78 | 18.75 | 68.37 | 0.583 (0.501, 0.664) |
| | | **B** | 2 | 13 | 2 | 1 | | | | | | | | | | |
| | | **C** | 1 | 63 | 105 | 15 | | | **High Density** | 65 | 132 | | | | | |
| | | **D** | 0 | 1 | 9 | 3 | | | | | | | | | | |
| Gemini-3.1 Pro | Without Mask | **A** | - | - | - | - | 0.852 | 56.28 | **Low Density** | 57 | 52 | 0.312 | 61.48 | 71.25 | 65.12 | 0.664 (0.589, 0.739) |
| | | **B** | 3 | 54 | 52 | 0 | | | | | | | | | | |
| | | **C** | 0 | 23 | 63 | 15 | | | **High Density** | 23 | 83 | | | | | |
| | | **D** | 0 | 0 | 1 | 4 | | | | | | | | | | |
| | With Mask | **A** | 0 | 1 | 0 | 0 | 0.871 | 53.02 | **Low Density** | 2 | 3 | 0.448 | 97.78 | 02.50 | 62.33 | 0.501 (0.421, 0.581) |
| | | **B** | 0 | 1 | 3 | 0 | | | | | | | | | | |
| | | **C** | 3 | 75 | 113 | 19 | | | **High Density** | 78 | 132 | | | | | |
| | | **D** | - | - | - | - | | | | | | | | | | |
| Sonnet-4.6 | Without Mask | **A** | 1 | 0 | 1 | 0 | 0.847 | 57.21 | **Low Density** | 28 | 9 | 0.530 | 93.33 | 35.00 | 71.63 | 0.642 (0.561, 0.722) |
| | | **B** | 2 | 25 | 8 | 0 | | | | | | | | | | |
| | | **C** | 0 | 48 | 92 | 14 | | | **High Density** | 52 | 126 | | | | | |
| | | **D** | 0 | 4 | 15 | 5 | | | | | | | | | | |
| | With Mask | **A** | 0 | 0 | 1 | 1 | 0.804 | 51.16 | **Low Density** | 11 | 6 | 0.464 | 95.56 | 13.75 | 65.12 | 0.547 (0.465, 0.628) |
| | | **B** | 3 | 8 | 4 | 0 | | | | | | | | | | |
| | | **C** | 0 | 59 | 93 | 9 | | | **High Density** | 69 | 129 | | | | | |
| | | **D** | 0 | 10 | 18 | 9 | | | | | | | | | | |
| Muse Spark | Without Mask | **A** | 2 | 15 | 7 | 0 | 0.772 | 51.16 | **Low Density** | 71 | 70 | 0.266 | 48.15 | 88.75 | 63.26 | 0.684 (0.613, 0.755) |
| | | **B** | 1 | 53 | 57 | 6 | | | | | | | | | | |
| | | **C** | 0 | 9 | 51 | 9 | | | **High Density** | 9 | 65 | | | | | |
| | | **D** | 0 | 0 | 1 | 4 | | | | | | | | | | |
| | With Mask | **A** | 0 | 4 | 1 | 0 | 0.824 | 56.74 | **Low Density** | 65 | 54 | 0.362 | 60.00 | 81.25 | 67.91 | 0.706 (0.635, 0.777) |
| | | **B** | 3 | 58 | 50 | 3 | | | | | | | | | | |
| | | **C** | 0 | 13 | 58 | 10 | | | **High Density** | 15 | 81 | | | | | |
| | | **D** | 0 | 2 | 7 | 6 | | | | | | | | | | |
| Radiologist 1 | - | **A** | 0 | 28 | 27 | 5 | 0.567 | 36.74 | **Low Density** | 53 | 68 | 0.120 | 49.63 | 66.25 | 55.81 | 0.579 (0.501, 0.658) |
| | | **B** | 2 | 23 | 32 | 4 | | | | | | | | | | |
| | | **C** | 1 | 25 | 54 | 8 | | | **High Density** | 27 | 67 | | | | | |
| | | **D** | 0 | 1 | 3 | 2 | | | | | | | | | | |
| | - | **A** | 2 | 0 | 0 | 0 | 0.886 | 67.91 | | 29 | 2 | 0.600 | 98.52 | 36.25 | 75.35 | |

| Radiologist 2 | | **B** | 1 | 26 | 2 | 0 | | | **Low Density** | | | | | | | 0.674 (0.594, 0.753) |
|---|---|---|---|---|---|---|---|---|---|---|---|---|---|---|---|---|
| | | **C** | 0 | 48 | 102 | 3 | | | **High Density** | 51 | 133 | | | | | |
| | | **D** | 0 | 3 | 12 | 16 | | | | | | | | | | |
| Radiologist 3 | | **A** | 3 | 9 | 0 | 0 | | | **Low Density** | 54 | 20 | | | | | |
| | - | **B** | 0 | 42 | 19 | 1 | 0.817 | 59.07 | | | | 0.604 | 85.18 | 67.50 | 78.60 | 0.763 (0.693, 0.834) |
| | | **C** | 0 | 22 | 66 | 2 | | | **High Density** | 26 | 115 | | | | | |
| | | **D** | 0 | 4 | 31 | 16 | | | | | | | | | | |

**Table 2.** Performance of MLLMS and radiologists in identifying ACR BI-RADS of lesions in digital mammography.

| Method | Condition | Predicted Class | True Class | | | | | Gwet's AC | Acc (%) | Predicted Class | True Class | | Gwet's AC | Sen (%) | Spec (%) | Acc (%) | AUC (95% CI) |
|---|---|---|---|---|---|---|---|---|---|---|---|---|---|---|---|---|---|
| | | | 1 | 2 | 3 | 4 | 5 | | | | Not Candid | Candid | | | | | |
| ChatGPT-5.2 | Without Mask | 1 | 2 | 0 | 1 | 2 | 1 | 0.569 | 35.35 | Not Candid | 53 | 41 | 0.349 | 68.70 | 63.09 | 66.51 | 0.659 (0.584, 0.734) |
| | | 2 | 5 | 10 | 18 | 22 | 4 | | | Candid | 31 | 90 | | | | | |
| | | 3 | 4 | 1 | 12 | 11 | 1 | | | | | | | | | | |
| | | 4 | 3 | 6 | 10 | 24 | 21 | | | | | | | | | | |
| | | 5 | 1 | 1 | 10 | 17 | 28 | | | | | | | | | | |
| | With Mask | 1 | 2 | 0 | 1 | 1 | 0 | 0.707 | 45.12 | Not Candid | 33 | 12 | 0.495 | 90.84 | 39.29 | 70.70 | 0.651 (0.572, 0.729) |
| | | 2 | 2 | 8 | 9 | 7 | 1 | | | Candid | 51 | 119 | | | | | |
| | | 3 | 2 | 2 | 7 | 3 | 0 | | | | | | | | | | |
| | | 4 | 9 | 6 | 28 | 43 | 17 | | | | | | | | | | |
| | | 5 | 0 | 2 | 6 | 22 | 37 | | | | | | | | | | |
| Gemini-3.1 Pro | Without Mask | 1 | 11 | 2 | 8 | 7 | 3 | 0.501 | 35.35 | Not Candid | 55 | 37 | 0.406 | 71.76 | 65.48 | 69.30 | 0.686 (0.612, 0.760) |
| | | 2 | 3 | 12 | 18 | 16 | 6 | | | Candid | 29 | 94 | | | | | |
| | | 3 | 0 | 0 | 1 | 5 | 0 | | | | | | | | | | |
| | | 4 | 0 | 1 | 5 | 9 | 3 | | | | | | | | | | |
| | | 5 | 1 | 3 | 19 | 39 | 43 | | | | | | | | | | |
| | With Mask | 1 | 0 | 0 | 0 | 0 | 0 | 0.587 | 31.16 | Not Candid | 1 | 0 | 0.435 | 100 | 01.19 | 61.39 | 0.506 (0.426, 0.585) |
| | | 2 | 0 | 0 | 0 | 0 | 0 | | | Candid | 83 | 131 | | | | | |
| | | 3 | 0 | 1 | 0 | 0 | 0 | | | | | | | | | | |
| | | 4 | 3 | 11 | 25 | 14 | 2 | | | | | | | | | | |
| | | 5 | 12 | 6 | 26 | 62 | 53 | | | | | | | | | | |
| Sonnet-4.6 | Without Mask | 1 | 1 | 0 | 0 | 1 | 0 | 0.539 | 36.74 | Not Candid | 24 | 17 | 0.390 | 87.02 | 28.57 | 64.17 | 0.578 (0.498 |
| | | 2 | 6 | 3 | 6 | 5 | 3 | | | | | | | | | | |
| | | 3 | 0 | 2 | 6 | 6 | 2 | | | Candid | 60 | 114 | | | | | |

| | | | | | | | | | | | | | | | | |
|---|---|---|---|---|---|---|---|---|---|---|---|---|---|---|---|---|
| | | **4** | 4 | 6 | 19 | 34 | 15 | | | | | | | | | , 0.658) |
| | | **5** | 4 | 7 | 20 | 30 | 35 | | | | | | | | | |
| | With Mask | **1** | 0 | 0 | 0 | 0 | 0 | 0.648 | 37.67 | **Not Candid** | 21 | 1 | 0.526 | 99.24 | 25.00 | 70.23 | 0.621 (0.541 , 0.701) |
| | | **2** | 0 | 1 | 1 | 0 | 0 | | | | | | | | | |
| | | **3** | 1 | 8 | 10 | 1 | 0 | | | **Candid** | 63 | 130 | | | | | |
| | | **4** | 7 | 5 | 24 | 18 | 3 | | | | | | | | | |
| | | **5** | 7 | 4 | 16 | 57 | 52 | | | | | | | | | |
| Muse Spark | Without Mask | **1** | 7 | 3 | 8 | 6 | 1 | 0.601 | 40.93 | **Not Candid** | 42 | 20 | 0.477 | 84.73 | 50.00 | 71.16 | 0.674 (0.597 , 0.750) |
| | | **2** | 1 | 7 | 9 | 7 | 1 | | | | | | | | | |
| | | **3** | 1 | 0 | 6 | 5 | 0 | | | **Candid** | 42 | 111 | | | | | |
| | | **4** | 4 | 6 | 17 | 28 | 13 | | | | | | | | | |
| | | **5** | 2 | 2 | 11 | 30 | 40 | | | | | | | | | |
| | With Mask | **1** | 3 | 3 | 4 | 1 | 0 | 0.681 | 40.93 | **Not Candid** | 38 | 7 | 0.575 | 94.66 | 45.24 | 75.35 | 0.699 (0.623 , 0.776) |
| | | **2** | 2 | 4 | 1 | 1 | 0 | | | | | | | | | |
| | | **3** | 1 | 7 | 13 | 5 | 0 | | | **Candid** | 46 | 124 | | | | | |
| | | **4** | 4 | 2 | 20 | 17 | 4 | | | | | | | | | |
| | | **5** | 5 | 2 | 13 | 52 | 51 | | | | | | | | | |
| Radiologist 1 | - | **1** | 5 | 1 | 1 | 0 | 1 | 0.763 | 60.93 | **Not Candid** | 76 | 24 | 0.708 | 81.68 | 90.48 | 85.12 | 0.861 (0.807 , 0.914) |
| | | **2** | 6 | 13 | 23 | 7 | 3 | | | | | | | | | |
| | | **3** | 2 | 1 | 24 | 10 | 3 | | | **Candid** | 8 | 107 | | | | | |
| | | **4** | 1 | 1 | 2 | 52 | 11 | | | | | | | | | |
| | | **5** | 1 | 2 | 1 | 7 | 37 | | | | | | | | | |
| Radiologist 2 | - | **1** | 12 | 2 | 2 | 1 | 0 | 0.876 | 67.44 | **Not Candid** | 61 | 7 | 0.743 | 94.66 | 72.62 | 86.05 | 0.836 (0.774 |
| | | **2** | 3 | 5 | 3 | 1 | 1 | | | | | | | | | |

| | | | | | | | | | | | | | | | | | |
|---|---|---|---|---|---|---|---|---|---|---|---|---|---|---|---|---|---|
| | | **3** | 0 | 7 | 27 | 4 | 0 | | | | | | | | | | , 0.898) |
| | | **4** | 0 | 4 | 19 | 64 | 17 | | | **Candid** | 23 | 124 | | | | | |
| | | **5** | 0 | 0 | 0 | 6 | 37 | | | | | | | | | | |
| Radiologist 3 | - | **1** | 0 | 0 | 0 | 0 | 0 | 0.839 | 56.74 | **Not Candid** | 79 | 17 | 0.801 | 87.02 | 94.05 | 89.77 | 0.905 (0.861, 0.950) |
| | | **2** | 9 | 7 | 3 | 1 | 0 | | | | | | | | | | |
| | | **3** | 6 | 11 | 43 | 14 | 2 | | | **Candid** | 5 | 114 | | | | | |
| | | **4** | 0 | 0 | 1 | 23 | 4 | | | | | | | | | | |
| | | **5** | 0 | 0 | 4 | 38 | 49 | | | | | | | | | | |

**Table 3.** Performance of MLLMS and radiologists in identifying ACR BI-RADS of lesions in contrast-enhanced mammography.

| Method | Condition | Predicted Class | True Class 1 | 2 | 3 | 4 | 5 | Gwet's AC | Acc (%) | Predicted Class | True Class Not Candid | Candid | Gwet's AC | Sen (%) | Spec (%) | Acc (%) | AUC (95% CI) |
|---|---|---|---|---|---|---|---|---|---|---|---|---|---|---|---|---|---|
| ChatGPT-5.2 | Without Mask | 1 | 0 | 4 | 2 | 2 | 1 | 0.694 | 46.98 | Not Candid | 37 | 21 | 0.568 | 85.52 | 52.86 | 74.88 | 0.692 (0.612, 0.772) |
| | | 2 | 1 | 18 | 5 | 7 | 4 | | | | | | | | | | |
| | | 3 | 0 | 4 | 3 | 3 | 4 | | | Candid | 33 | 124 | | | | | |
| | | 4 | 1 | 16 | 10 | 20 | 29 | | | | | | | | | | |
| | | 5 | 0 | 5 | 1 | 15 | 60 | | | | | | | | | | |
| | With Mask | 1 | 0 | 3 | 0 | 0 | 0 | 0.735 | 51.16 | Not Candid | 38 | 15 | 0.630 | 89.65 | 54.29 | 78.14 | 0.720 (0.641, 0.799) |
| | | 2 | 1 | 16 | 8 | 3 | 7 | | | | | | | | | | |
| | | 3 | 0 | 8 | 2 | 5 | 0 | | | Candid | 32 | 130 | | | | | |
| | | 4 | 0 | 15 | 7 | 24 | 23 | | | | | | | | | | |
| | | 5 | 1 | 5 | 4 | 15 | 68 | | | | | | | | | | |
| Gemini-3.1 Pro | Without Mask | 1 | 0 | 16 | 5 | 5 | 5 | 0.503 | 40.00 | Not Candid | 43 | 39 | 0.435 | 73.10 | 61.43 | 69.30 | 0.673 (0.594, 0.751) |
| | | 2 | 1 | 15 | 6 | 14 | 15 | | | | | | | | | | |
| | | 3 | 0 | 0 | 0 | 0 | 0 | | | Candid | 27 | 106 | | | | | |
| | | 4 | 1 | 10 | 4 | 17 | 24 | | | | | | | | | | |
| | | 5 | 0 | 6 | 6 | 11 | 54 | | | | | | | | | | |
| | With Mask | 1 | 0 | 2 | 0 | 0 | 0 | 0.714 | 42.33 | Not Candid | 19 | 4 | 0.613 | 97.24 | 27.14 | 74.42 | 0.622 (0.537, 0.707) |
| | | 2 | 0 | 9 | 7 | 2 | 1 | | | | | | | | | | |
| | | 3 | 0 | 1 | 0 | 0 | 1 | | | Candid | 51 | 141 | | | | | |
| | | 4 | 1 | 28 | 11 | 30 | 44 | | | | | | | | | | |
| | | 5 | 1 | 7 | 3 | 15 | 52 | | | | | | | | | | |
| Sonnet-4.6 | | 1 | 0 | 0 | 0 | 0 | 0 | 0.722 | | | 22 | 7 | 0.604 | | | | |
| | Without Mask | **2** | 0 | 9 | 4 | 3 | 1 | | 45.12 | **Not Candid** | | | | 95.17 | 31.43 | 74.42 | 0.633 (0.549, 0.717) |
| | | **3** | 0 | 6 | 3 | 3 | 0 | | | **Candid** | 48 | 138 | | | | | |
| | | **4** | 2 | 22 | 8 | 16 | 28 | | | | | | | | | | |
| | | **5** | 0 | 10 | 6 | 25 | 69 | | | | | | | | | | |
| | With Mask | **1** | 0 | 0 | 0 | 0 | 0 | 0.754 | 46.51 | **Not Candid** | 15 | 1 | 0.617 | 99.31 | 21.43 | 73.95 | 0.604 (0.518, 0.689) |
| | | **2** | 0 | 1 | 0 | 0 | 0 | | | | | | | | | | |
| | | **3** | 0 | 10 | 4 | 1 | 0 | | | **Candid** | 55 | 144 | | | | | |
| | | **4** | 2 | 26 | 11 | 7 | 10 | | | | | | | | | | |
| | | **5** | 0 | 10 | 6 | 39 | 88 | | | | | | | | | | |
| Muse Spark | Without Mask | **1** | 0 | 19 | 5 | 4 | 2 | 0.707 | 48.84 | **Not Candid** | 54 | 24 | 0.661 | 83.45 | 77.14 | 81.39 | 0.803 (0.736, 0.870) |
| | | **2** | 2 | 20 | 6 | 6 | 6 | | | | | | | | | | |
| | | **3** | 0 | 2 | 0 | 3 | 3 | | | **Candid** | 16 | 121 | | | | | |
| | | **4** | 0 | 4 | 8 | 16 | 18 | | | | | | | | | | |
| | | **5** | 0 | 2 | 2 | 18 | 69 | | | | | | | | | | |
| | With Mask | **1** | 0 | 8 | 0 | 4 | 1 | 0.787 | 55.81 | **Not Candid** | 46 | 13 | 0.730 | 91.03 | 65.71 | 82.79 | 0.784 (0.711, 0.857) |
| | | **2** | 2 | 20 | 7 | 3 | 0 | | | | | | | | | | |
| | | **3** | 0 | 7 | 2 | 4 | 1 | | | **Candid** | 24 | 132 | | | | | |
| | | **4** | 0 | 7 | 8 | 16 | 14 | | | | | | | | | | |
| | | **5** | 0 | 5 | 4 | 20 | 82 | | | | | | | | | | |
| Radiologist 1 | - | **1** | 0 | 0 | 0 | 1 | 0 | 0.923 | 79.53 | **Not Candid** | 67 | 16 | 0.837 | 88.96 | 95.71 | 91.16 | 0.923 (0.883, 0.964) |
| | | **2** | 2 | 41 | 5 | 2 | 1 | | | | | | | | | | |
| | | **3** | 0 | 6 | 13 | 10 | 2 | | | **Candid** | 3 | 129 | | | | | |
| | | **4** | 0 | 0 | 3 | 28 | 6 | | | | | | | | | | |
| | | **5** | 0 | 0 | 0 | 6 | 89 | | | | | | | | | | |
| Radiologist 2 | - | **1** | 0 | 0 | 0 | 0 | 0 | 0.939 | 82.79 | **Not Candid** | 58 | 5 | 0.862 | 96.55 | 82.86 | 92.09 | 0.897 (0.842, 0.952) |
| | | **2** | 1 | 35 | 0 | 0 | 0 | | | | | | | | | | |
| | | **3** | 1 | 7 | 14 | 5 | 0 | | | **Candid** | 12 | 140 | | | | | |
| | | **4** | 0 | 5 | 7 | 39 | 8 | | | | | | | | | | |
| | | **5** | 0 | 0 | 0 | 3 | 90 | | | | | | | | | | |
| Radiologist 3 | - | **1** | 0 | 0 | 0 | 0 | 0 | 0.872 | 62.33 | **Not Candid** | 66 | 24 | 0.755 | 83.45 | 94.29 | 86.98 | 0.889 (0.841, 0.937) |
| | | **2** | 1 | 16 | 0 | 2 | 1 | | | | | | | | | | |
| | | **3** | 1 | 30 | 18 | 18 | 3 | | | **Candid** | 4 | 121 | | | | | |
| | | **4** | 0 | 1 | 1 | 14 | 8 | | | | | | | | | | |
| | | **5** | 0 | 0 | 2 | 13 | 86 | | | | | | | | | | |

**Table 4.** Performance of MLLMS and radiologists in identifying benign and malignant lesions in digital and contrast-enhanced mammography.

| Modality | Method | Condition | Predicted Class | True Class | | Gwet's AC | Sen (%) | Spec (%) | Acc (%) | AUC (95% CI) |
|---|---|---|---|---|---|---|---|---|---|---|
| | | | | Benign | Malignant | | | | | |
| Digital Mammography (DM) | ChatGPT-5.2 | Without Mask | **Benign** | 52 | 43 | 0.321 | 67.18 | 61.90 | 65.12 | 0.667 (0.592, 742) |
| | | | **Malignant** | 32 | 88 | | | | | |
| | | With Mask | **Benign** | 35 | 10 | 0.527 | 92.37 | 41.67 | 72.56 | 0.791 (0.729, 0.854) |
| | | | **Malignant** | 49 | 121 | | | | | |
| | Gemini-3.1 Pro | Without Mask | **Benign** | 58 | 35 | 0.450 | 73.28 | 69.05 | 71.63 | 0.703 (0.630, 0.775) |
| | | | **Malignant** | 26 | 96 | | | | | |
| | | With Mask | **Benign** | 42 | 8 | 0.593 | 93.89 | 50.00 | 76.74 | 0.759 (0.688, 0.829) |
| | | | **Malignant** | 42 | 123 | | | | | |
| | Sonnet-4.6 | Without Mask | **Benign** | 30 | 22 | 0.377 | 83.21 | 35.71 | 64.65 | 0.656 (0.582, 0.731) |
| | | | **Malignant** | 54 | 109 | | | | | |
| | | With Mask | **Benign** | 37 | 2 | 0.615 | 98.47 | 44.05 | 77.21 | 0.874 (0.823, 0.925) |
| | | | **Malignant** | 47 | 129 | | | | | |
| | Muse Spark | Without Mask | **Benign** | 45 | 22 | 0.479 | 83.21 | 53.57 | 71.63 | 0.728 (0.658, 0.798) |
| | | | **Malignant** | 39 | 109 | | | | | |
| | | With Mask | **Benign** | 48 | 10 | 0.616 | 92.37 | 57.14 | 78.60 | 0.834 (0.774, 0.894) |
| | | | **Malignant** | 36 | 121 | | | | | |
| | Radiologist 1 | - | **Benign** | 78 | 72 | 0.280 | 45.04 | 92.86 | 63.72 | 0.805 (0.742, 0.868) |
| | | | **Malignant** | 6 | 59 | | | | | |
| | Radiologist 2 | - | **Benign** | 73 | 26 | 0.663 | 80.15 | 86.90 | 82.79 | 0.865 (0.815, 0.915) |
| | | | **Malignant** | 11 | 105 | | | | | |
| | Radiologist 3 | - | **Benign** | 77 | 32 | 0.641 | 75.57 | 91.67 | 81.86 | 0.883 (0.833, 0.932) |
| | | | **Malignant** | 7 | 99 | | | | | |
| Subtracted-enhanced Mammography (CEM) | ChatGPT-5.2 | Without Mask | **Benign** | 45 | 16 | 0.537 | 87.79 | 53.57 | 74.42 | 0.799 (0.739, 0.859) |
| | | | **Malignant** | 39 | 115 | | | | | |
| | | With Mask | **Benign** | 44 | 11 | 0.578 | 91.6 | 52.38 | 76.28 | 0.789 (0.726, 0.852) |
| | | | **Malignant** | 40 | 120 | | | | | |
| | Gemini-3.1 Pro | Without Mask | **Benign** | 58 | 43 | 0.370 | 67.18 | 69.05 | 67.91 | 0.720 (0.649, 0.791) |
| | | | **Malignant** | 26 | 88 | | | | | |
| | | With Mask | **Benign** | 43 | 13 | 0.552 | 90.08 | 51.19 | 74.88 | 0.761 (0.694, 0.828) |
| | | | **Malignant** | 41 | 118 | | | | | |
| | Sonnet-4.6 | Without Mask | **Benign** | 31 | 8 | 0.520 | 93.89 | 36.9 | 71.63 | 0.752 (0.685, 0.818) |
| | | | **Malignant** | 53 | 123 | | | | | |
| | | With Mask | **Benign** | 25 | 0 | 0.558 | 100 | 29.76 | 72.56 | 0.874 (0.826, 0.922) |
| | | | **Malignant** | 59 | 131 | | | | | |
| | Muse Spark | Without Mask | **Benign** | 66 | 31 | 0.555 | 76.34 | 78.57 | 77.21 | 0.826 (0.768, 0.885) |
| | | | **Malignant** | 18 | 100 | | | | | |
| | | With Mask | **Benign** | 58 | 13 | 0.663 | 90.08 | 69.05 | 81.86 | 0.875 (0.826, 0.923) |
| | | | **Malignant** | 26 | 118 | | | | | |
| | Radiologist 1 | - | **Benign** | 78 | 33 | 0.64 | 74.81 | 92.86 | 81.86 | 0.886 (0.839, 0.933) |
| | | | **Malignant** | 6 | 98 | | | | | |
| | Radiologist 2 | - | **Benign** | 70 | 10 | 0.789 | 92.37 | 83.33 | 88.84 | 0.941 (0.906, 0.977) |
| | | | **Malignant** | 14 | 121 | | | | | |
| | Radiologist 3 | - | **Benign** | 76 | 24 | 0.708 | 81.68 | 90.48 | 85.12 | 0.901 (0.856, 0.947) |
| | | | **Malignant** | 8 | 107 | | | | | |

**Table 5.** Comparison of the present study with previous studies regarding (1) digital mammography (DM) evaluation, (2) contrast-enhanced mammography (CEM) evaluation, (3) four-class breast density classification, (4) two-class (high vs. low) breast density classification, (5) BI-RADS score classification, (6) biopsy candidacy determination, (7) malignancy determination, (8) comparison with radiologists, and (9) mask effect evaluation.

| **Study** | **Model** | **DM** | **CEM** | **Density classification** | **High vs. low density classification** | **BI-RADS score classification** | **Biopsy candidacy determination** | **Malignancy determination** | **Comparison with radiologist** | **Mask effect evaluation** |
|---|---|---|---|---|---|---|---|---|---|---|
| Haver et al., 2024 (23) | ChatGPT-4V | ✓ | ✗ | ✗ | ✗ | ✓ | ✗ | ✗ | ✗ | ✗ |
| Ra et al., 2025 (27) | LLaMA2 | ✓ | ✗ | ✗ | ✗ | ✗ | ✗ | ✓ (only based on BI-RADS) | ✗ | ✗ |
| Sanli et al., 2025 (18) | XrayGPT | ✓ | ✗ | ✗ | ✓ | ✓ (only 4 vs. 5) | ✗ | ✗ | ✗ | ✗ |
| Nguyen et al., 2025 (22) | ChatGPT-4, ChatGPT-4o | ✓ | ✗ | ✗ | ✗ | ✓ | ✗ | ✗ | ✗ | ✗ |
| Karahan et al., 2025 (20) | ChatGPT-4o, Claude 3.5 | ✓ | ✗ | ✓ | ✗ | ✓ | ✗ | ✗ | ✗ | ✗ |
| Zhu et al., 2025 (21) | LLaVA-Mammo, miniGPT-4, InstructBLIP, BlIP-2, LLaVA-NeXT, InternVL3, Qwen2.5-VL, LLaVA-Med, RadFM, Med-Flamingo, MedVlnT-TD, MedDr, MedGemma | ✓ | ✗ | ✓ | ✗ | ✓ | ✗ | ✓ | ✗ | ✗ |
| Küskün et al., 2026 (24) | ChatGPT-4o | ✓ | ✗ | ✗ | ✗ | ✓ | ✓ | ✗ | ✓ (three radiologists) only for BI-RADS Score | ✗ |
| Li et al., 2026 (19) | ChatGPT-5, ChatGPT-4o | ✓ | ✗ | ✓ | ✗ | ✓ | ✗ | ✓ | ✓ (one radiologist) only for malignancy | ✗ |
| Chen et al., 2026 (26) | Mammo-CLIP | ✓ | ✗ | ✗ | ✗ | ✗ | ✗ | ✓ | ✗ | ✗ |
| Alarifi et al., 2026 (25) | ChatGPT-5.2 Thinking | ✓ | ✗ | ✗ | ✗ | ✗ | ✗ | ✓ | ✓ (one radiologist) only for BI-RADS Score | ✗ |
| Our study | ChatGPT-5.2, Gemini-3.1 Pro, Sonnet-4.6, Muse Spark | ✓ | ✓ | ✓ | ✓ | ✓ | ✓ | ✓ | ✓ (three radiologists) for density, BI-RADS Score, biopsy candidacy) | ✓ |

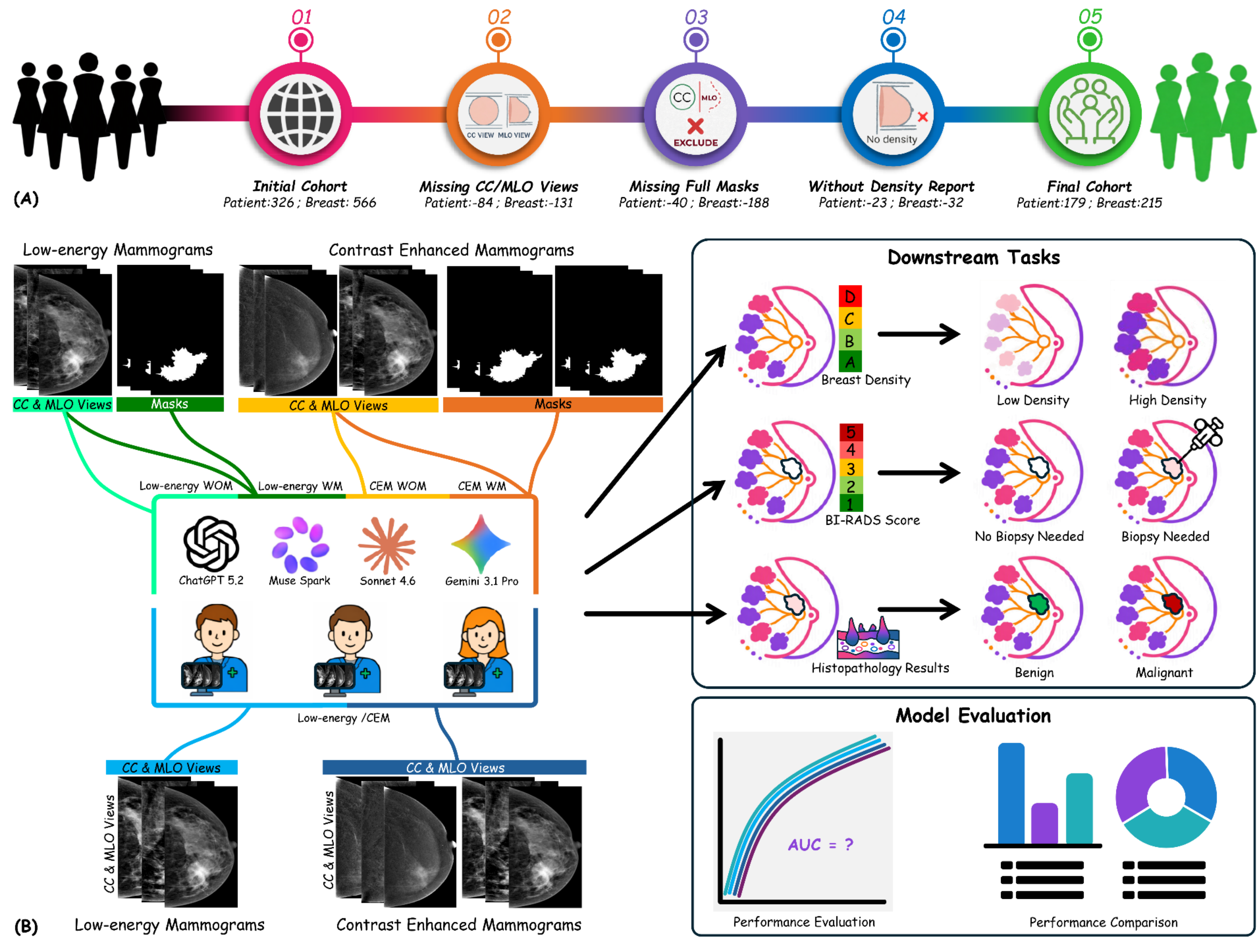
01
02
03
04
05
Initial Cohort
Patient:326 ; Breast: 566
Missing CC/MLO Views
Patient:-84 ; Breast:-131
Missing Full Masks
Patient:-40 ; Breast:-188
Without Density Report
Patient:-23 ; Breast:-32
Final Cohort
Patient:179 ; Breast:215
(A)
Low-energy Mammograms
Contrast Enhanced Mammograms
CC & MLO Views
Masks
Low-energy WOM
Low-energy WM
CEM WOM
CEM WM
ChatGPT 5.2
Muse Spark
Sonnet 4.6
Gemini 3.1 Pro
Low-energy /CEM
Downstream Tasks
Breast Density
Low Density
High Density
BI-RADS Score
No Biopsy Needed
Biopsy Needed
Histopathology Results
Benign
Malignant
Model Evaluation
AUC = ?
Performance Evaluation
Performance Comparison
(B)

**Fig. 1. (A)** Flow diagram of patient selection in this study. After a three-step exclusion process, 179 patients, including 215 breasts, were included in the final analytical cohort. **(B)** Workflow diagram used in this study. Four MLLMs (ChatGPT-5.2, Muse Spark, Sonnet-4.6, and Gemini-3.1 Pro) along with three radiologists interpreted the images for four downstream tasks: predicting (1) ACR breast density category, (2) ACR BI-RADS score, (3) candidacy for biopsy, and (4) benign versus malignant breast lesions.

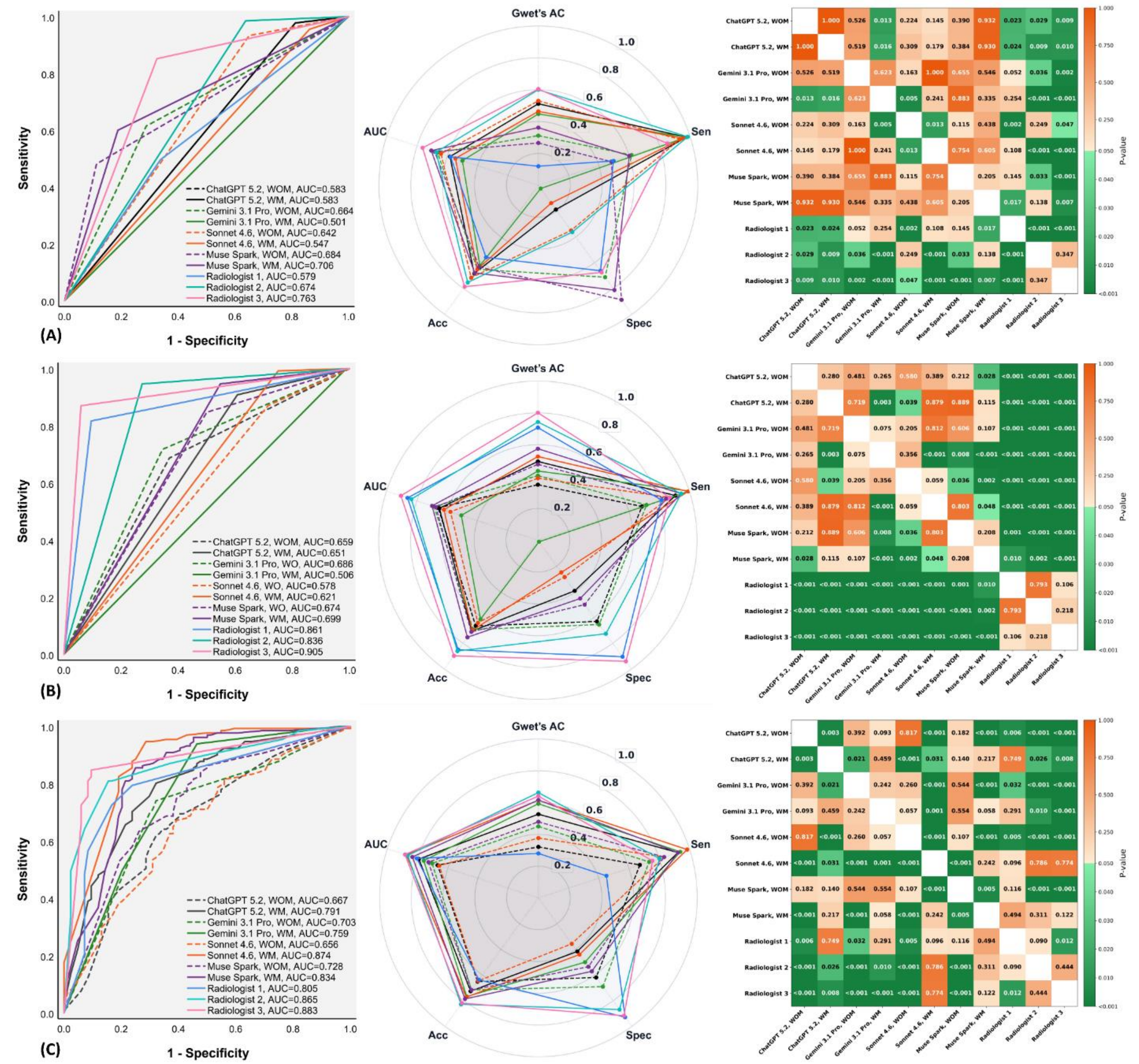

Sensitivity
1 - Specificity
(A)
ChatGPT 5.2, WOM, AUC=0.583
ChatGPT 5.2, WM, AUC=0.583
Gemini 3.1 Pro, WOM, AUC=0.664
Gemini 3.1 Pro, WM, AUC=0.501
Sonnet 4.6, WOM, AUC=0.642
Sonnet 4.6, WM, AUC=0.547
Muse Spark, WOM, AUC=0.684
Muse Spark, WM, AUC=0.706
Radiologist 1, AUC=0.579
Radiologist 2, AUC=0.674
Radiologist 3, AUC=0.763
Gwet's AC
Sen
Spec
Acc
AUC
P-value
(B)
ChatGPT 5.2, WOM, AUC=0.659
ChatGPT 5.2, WM, AUC=0.651
Gemini 3.1 Pro, WO, AUC=0.686
Gemini 3.1 Pro, WM, AUC=0.506
Sonnet 4.6, WO, AUC=0.578
Sonnet 4.6, WM, AUC=0.621
Muse Spark, WO, AUC=0.674
Muse Spark, WM, AUC=0.699
Radiologist 1, AUC=0.861
Radiologist 2, AUC=0.836
Radiologist 3, AUC=0.905
(C)
ChatGPT 5.2, WOM, AUC=0.667
ChatGPT 5.2, WM, AUC=0.791
Gemini 3.1 Pro, WOM, AUC=0.703
Gemini 3.1 Pro, WM, AUC=0.759
Sonnet 4.6, WOM, AUC=0.656
Sonnet 4.6, WM, AUC=0.874
Muse Spark, WOM, AUC=0.728
Muse Spark, WM, AUC=0.834
Radiologist 1, AUC=0.805
Radiologist 2, AUC=0.865
Radiologist 3, AUC=0.883

**Fig. 2.** ROC curve (left), radar plot (middle), and comparison heat map (right) of four MLLMs and three radiologists in identifying low and high breast density (A), breast lesion candidates for biopsy (B), and benign and malignant breast lesions (C) in low-energy mammograms. WOM, without mask; WM, with mask; Sen, sensitivity; Spec, specificity; ACC, accuracy; AUC, area under the ROC curve.

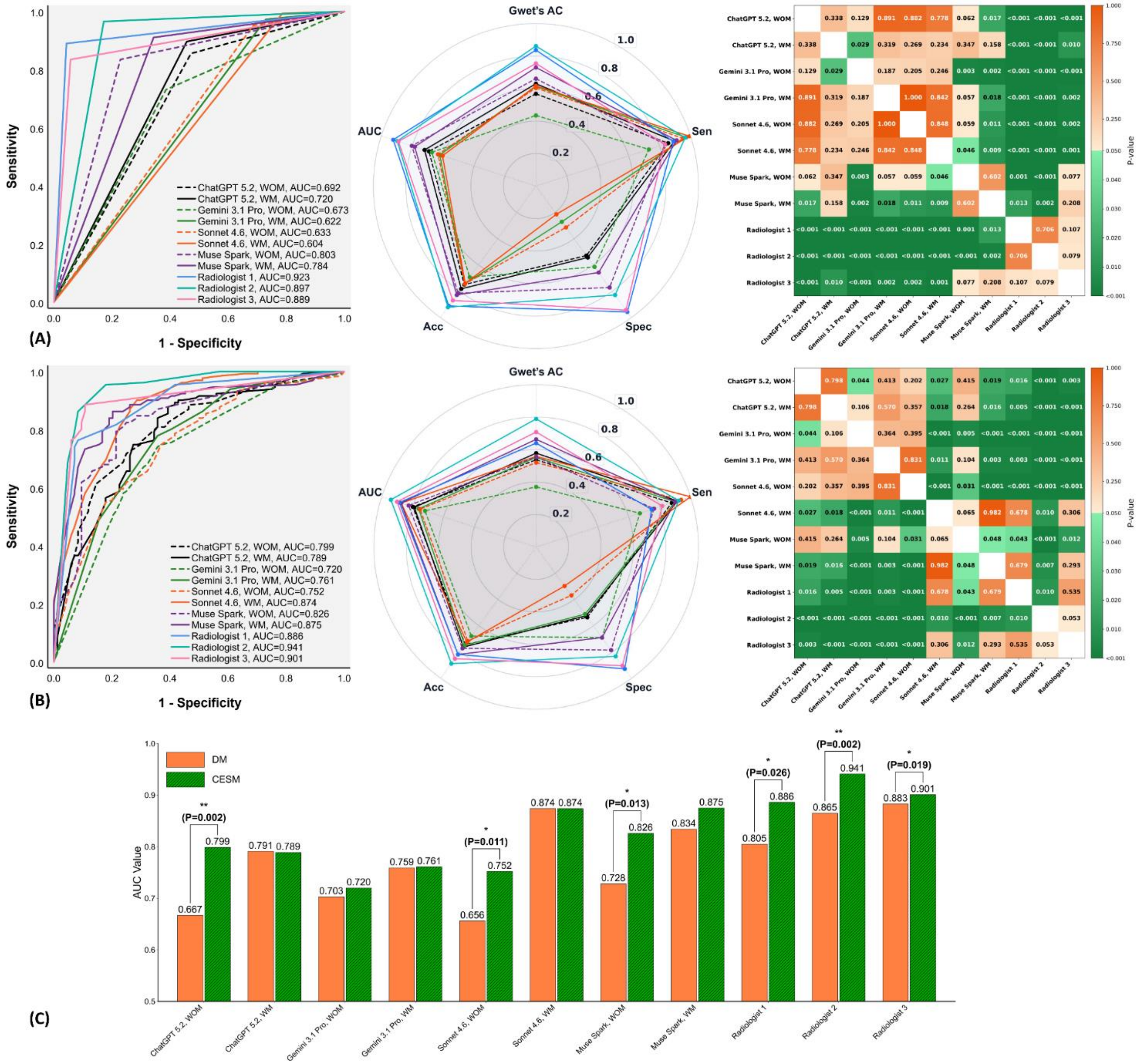

(A)
Sensitivity
1 - Specificity
ChatGPT 5.2, WOM, AUC=0.692
ChatGPT 5.2, WM, AUC=0.720
Gemini 3.1 Pro, WOM, AUC=0.673
Gemini 3.1 Pro, WM, AUC=0.622
Sonnet 4.6, WOM, AUC=0.633
Sonnet 4.6, WM, AUC=0.604
Muse Spark, WOM, AUC=0.803
Muse Spark, WM, AUC=0.784
Radiologist 1, AUC=0.923
Radiologist 2, AUC=0.897
Radiologist 3, AUC=0.889
Gwet's AC
AUC
Sen
Spec
Acc
P-value
(B)
Sensitivity
1 - Specificity
ChatGPT 5.2, WOM, AUC=0.799
ChatGPT 5.2, WM, AUC=0.789
Gemini 3.1 Pro, WOM, AUC=0.720
Gemini 3.1 Pro, WM, AUC=0.761
Sonnet 4.6, WOM, AUC=0.752
Sonnet 4.6, WM, AUC=0.874
Muse Spark, WOM, AUC=0.826
Muse Spark, WM, AUC=0.875
Radiologist 1, AUC=0.886
Radiologist 2, AUC=0.941
Radiologist 3, AUC=0.901
Gwet's AC
AUC
Sen
Spec
Acc
P-value
(C)
DM
CESM
AUC Value
** (P=0.002)
* (P=0.011)
* (P=0.013)
* (P=0.026)
** (P=0.002)
* (P=0.019)
ChatGPT 5.2, WOM
ChatGPT 5.2, WM
Gemini 3.1 Pro, WOM
Gemini 3.1 Pro, WM
Sonnet 4.6, WOM
Sonnet 4.6, WM
Muse Spark, WOM
Muse Spark, WM
Radiologist 1
Radiologist 2
Radiologist 3

**Fig. 3.** ROC curve (left), radar plot (middle), and comparison heat map (right) of four MLLMs and three radiologists in identifying breast lesion candidates for biopsy (A) and benign and malignant breast lesions (B) in contrast-enhanced mammography (CEM). A bar chart of AUC values of four MLLMs and three radiologists in identifying benign and malignant breast lesions for digital mammography (DM) and CEM is shown in (C). WOM, without mask; WM, with mask; Sen, sensitivity; Spec, specificity; ACC, accuracy; AUC, area under the ROC curve.

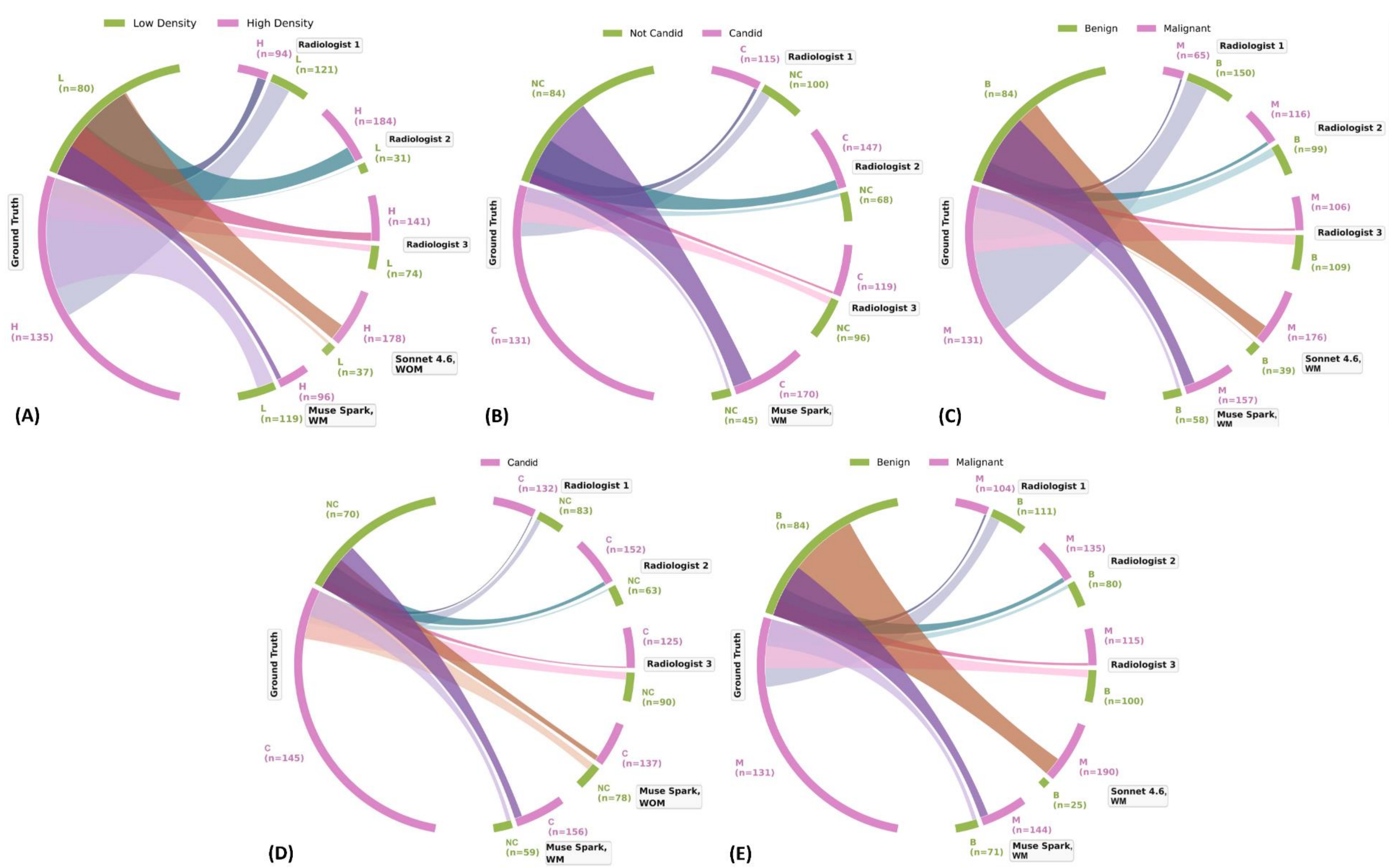

**Fig. 4.** Sankey flow diagram of the best MLLM(s) and three radiologists in identifying low and high breast density (A), breast lesion candidates for biopsy (B), and benign and malignant breast lesions (C) in low-energy mammograms. Sankey flow diagrams for identifying breast lesion candidates for biopsy and benign and malignant breast lesions in subtracted contrast-enhanced spectral

mammograms are depicted in (D) and (E), respectively. Only misdiagnosed cases are highlighted for better visualization. Type I (false-positive) and Type II (false-negative) errors are shown by dark and soft colors, respectively. WOM, without mask; WM, with mask.

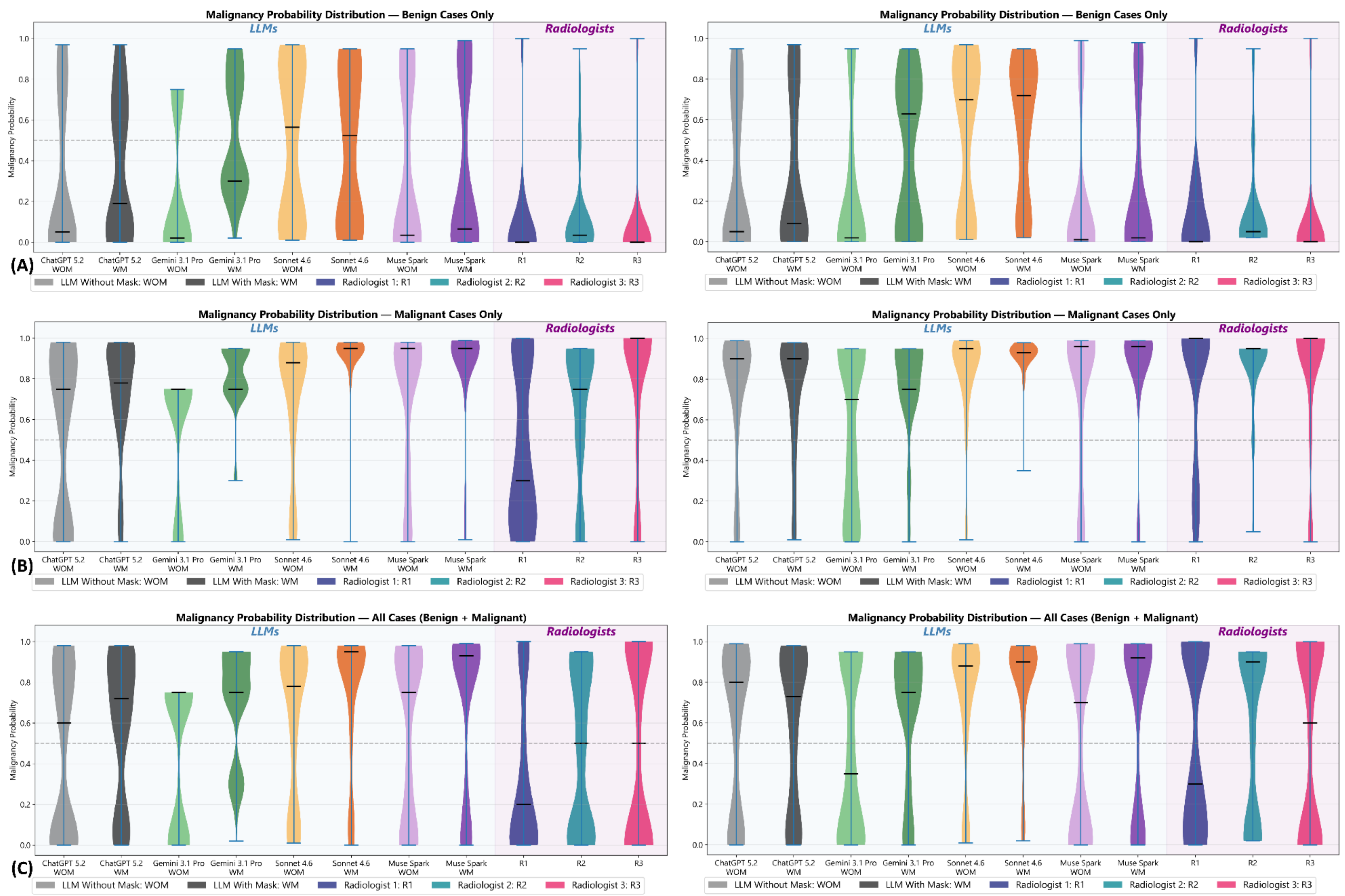

**Fig. 5.** Malignancy probability distribution of four MLLMs in two different approaches and three radiologists for benign (A), malignant (B), and all (C) breast lesions in low-energy (first column) and subtracted contrast-enhanced spectral (second column) mammograms.

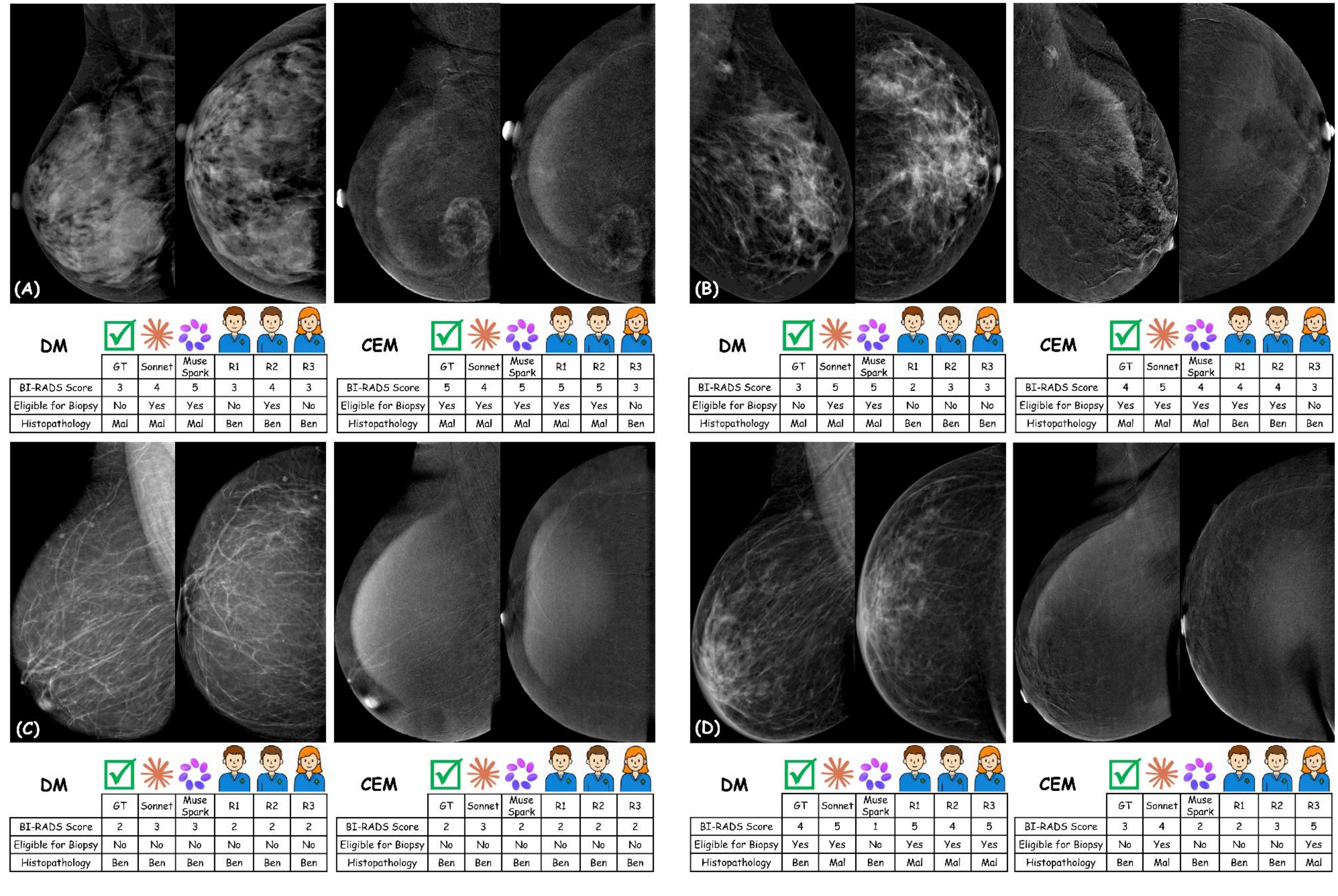

(A) DM

| | GT | Sonnet | Muse Spark | R1 | R2 | R3 |
|---|---|---|---|---|---|---|
| BI-RADS Score | 3 | 4 | 5 | 3 | 4 | 3 |
| Eligible for Biopsy | No | Yes | Yes | No | Yes | No |
| Histopathology | Mal | Mal | Mal | Ben | Ben | Ben |

(A) CEM

| | GT | Sonnet | Muse Spark | R1 | R2 | R3 |
|---|---|---|---|---|---|---|
| BI-RADS Score | 5 | 4 | 5 | 5 | 5 | 3 |
| Eligible for Biopsy | Yes | Yes | Yes | Yes | Yes | No |
| Histopathology | Mal | Mal | Mal | Mal | Mal | Ben |

(B) DM

| | GT | Sonnet | Muse Spark | R1 | R2 | R3 |
|---|---|---|---|---|---|---|
| BI-RADS Score | 3 | 5 | 5 | 2 | 3 | 3 |
| Eligible for Biopsy | No | Yes | Yes | No | No | No |
| Histopathology | Mal | Mal | Mal | Ben | Ben | Ben |

(B) CEM

| | GT | Sonnet | Muse Spark | R1 | R2 | R3 |
|---|---|---|---|---|---|---|
| BI-RADS Score | 4 | 5 | 4 | 4 | 4 | 3 |
| Eligible for Biopsy | Yes | Yes | Yes | Yes | Yes | No |
| Histopathology | Mal | Mal | Mal | Ben | Ben | Ben |

(C) DM

| | GT | Sonnet | Muse Spark | R1 | R2 | R3 |
|---|---|---|---|---|---|---|
| BI-RADS Score | 2 | 3 | 3 | 2 | 2 | 2 |
| Eligible for Biopsy | No | No | No | No | No | No |
| Histopathology | Ben | Ben | Ben | Ben | Ben | Ben |

(C) CEM

| | GT | Sonnet | Muse Spark | R1 | R2 | R3 |
|---|---|---|---|---|---|---|
| BI-RADS Score | 2 | 3 | 2 | 2 | 2 | 2 |
| Eligible for Biopsy | No | No | No | No | No | No |
| Histopathology | Ben | Ben | Ben | Ben | Ben | Ben |

(D) DM

| | GT | Sonnet | Muse Spark | R1 | R2 | R3 |
|---|---|---|---|---|---|---|
| BI-RADS Score | 4 | 5 | 1 | 5 | 4 | 5 |
| Eligible for Biopsy | Yes | Yes | No | Yes | Yes | Yes |
| Histopathology | Ben | Mal | Ben | Mal | Mal | Mal |

(D) CEM

| | GT | Sonnet | Muse Spark | R1 | R2 | R3 |
|---|---|---|---|---|---|---|
| BI-RADS Score | 3 | 4 | 2 | 2 | 3 | 5 |
| Eligible for Biopsy | No | Yes | No | No | No | Yes |
| Histopathology | Ben | Mal | Ben | Ben | Ben | Mal |

**Fig. 6.** Results of three radiologists (R1, R2, R3) and the two best MLLMs, Sonnet-4.6 and Muse Spark, operating with lesion masks, in ACR BI-RADS scoring, biopsy candidacy assessment, and malignancy prediction for malignant **(A-B)** and benign **(C-D)** breast lesions. Ben, Benign; Mal, Malignant.